\documentclass[10pt,twocolumn,letterpaper]{article}
\usepackage{cvpr}
\usepackage{times}
\usepackage{graphicx}
\usepackage{amsmath}
\usepackage{amssymb}
\usepackage{booktabs}
\cvprfinalcopy

\usepackage[pagebackref=true,breaklinks=true,colorlinks,bookmarks=false]{hyperref}

\begin{document}

\title{WenLan: Bridging Vision and Language by Large-Scale \\ Multi-Modal Pre-Training }

\author{
Yuqi Huo$^2$ Manli Zhang$^2$ Guangzhen Liu$^2$ Haoyu Lu$^1$ Yizhao Gao$^1$ Guoxing Yang$^1$ Jingyuan Wen$^1$\\
Heng Zhang$^1$ Baogui Xu$^1$ Weihao Zheng$^2$ Zongzheng Xi$^2$ Yueqian Yang$^1$ Anwen Hu$^2$ Jinming Zhao$^2$ \\
Ruichen Li$^2$ Yida Zhao$^2$ Liang Zhang$^2$ Yuqing Song$^2$ Xin Hong$^3$ Wanqing Cui$^3$ Danyang Hou$^3$\\
Yingyan Li$^3$ Junyi Li$^1$ Peiyu Liu$^1$ Zheng Gong$^1$  Chuhao Jin$^1$ Yuchong Sun$^1$ Shizhe Chen$^2$\\
Zhiwu Lu$^{1}\thanks{Co-corresponding authors.}$~~~Zhicheng Dou$^1$ Qin Jin$^2$ Yanyan Lan$^3$ Wayne Xin Zhao$^1$ Ruihua Song$^{1 *}$ Ji-Rong Wen$^{1 *}$\\
$^1$Gaoling School of Artificial Intelligence, Renmin University of China, Beijing, China\\
$^2$School of Information, Renmin University of China, Beijing, China\\
$^3$Institute of Computing Technology, Chinese Academy of Sciences, Beijing, China\\
{\tt\small \{luzhiwu, rsong, jrwen\}@ruc.edu.cn}
}

\maketitle

\begin{abstract}
Multi-modal pre-training models have been intensively explored to bridge vision and language in recent years. However, most of them explicitly model the cross-modal interaction between image-text pairs, by assuming that there exists strong semantic correlation between the text and image modalities. Since this strong assumption is often invalid in real-world scenarios, we choose to implicitly model the cross-modal correlation for large-scale multi-modal pre-training, which is the focus of the Chinese project `WenLan' led by our team. Specifically, with the weak correlation assumption over image-text pairs, we propose a two-tower pre-training model called BriVL within the cross-modal contrastive learning framework. Unlike OpenAI CLIP that adopts a simple contrastive learning method, we devise a more advanced algorithm by adapting the latest method MoCo into the cross-modal scenario. By building a large queue-based dictionary, our BriVL can incorporate more negative samples in limited GPU resources. We further construct a large Chinese multi-source image-text dataset called RUC-CAS-WenLan for pre-training our BriVL model. Extensive experiments demonstrate that the pre-trained BriVL model outperforms both UNITER and OpenAI CLIP on various downstream tasks.
\end{abstract}

\section{Introduction}

\begin{figure}[t]
    \centering
    \includegraphics[width=0.47\textwidth]{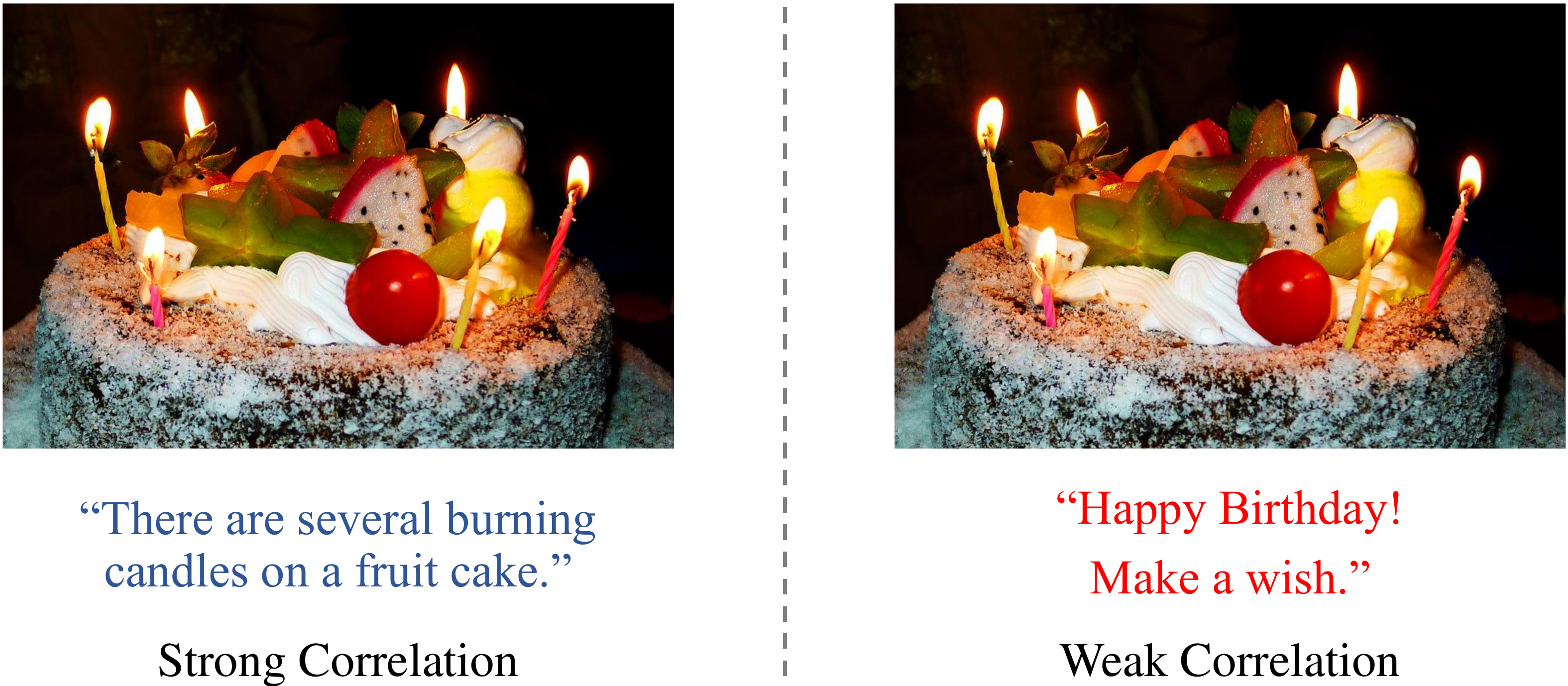}
    \caption{
    Example of strong correlation assumption versus weak correlation assumption over image-text pairs. Note that the strong correlation assumption widely used in many multi-model pre-training models is often invalid in real-world scenarios. }
    \label{fig:intro}
    \vspace{-0.05in}
\end{figure}

In recent years, pre-training models have become topical in natural language processing (NLP). A number of pre-training language models such as BERT~\cite{devlin2018bert,liu2019roberta,Lan2020ALBERT} and GPT~\cite{radford2018improving,radford2019language,brown2020language} have achieved significant improvements on various downstream NLP tasks. With the release of GPT-3~\cite{brown2020language} (i.e., the latest large-scale language model of OpenAI), pre-training language models~\cite{radford2021learning,roberts2020much,guu2020realm} have now drawn the most attention of the NLP community.

Compared with text understanding in the single-modal scenario, understanding multiple modalities is more attractive and has a broader rang of application scenarios. In fact, with the success of pre-training models in NLP~\cite{dai2015semi,peters2018deep,howard2018universal,raffel2019exploring}, they have recently been extended to understand the text and the image simultaneously, that is, multi-modal pre-training models have been intensively explored to bridge vision and language in the last two years. Particularly, in January 2021, OpenAI released a multi-modal version of GPT-3~\cite{brown2020language} called DALL·E~\cite{aditya2021dall}, demonstrating its excellent text-to-image generation capability. This clearly declares the power of multi-modal pre-training, and also encourages researchers to explore the potential of large-scale multi-modal pre-training in the vision+language area.

Along this line of research, our team started a Chinese project called `WenLan' on large-scale multi-modal pre-training since September 2020, and released the first version to demonstrate its understanding ability on the Chinese multi-modal data. At this moment, our released model presents the strong image-text retrieval ability as well as the impressive commonsense understanding ability.

As we have mentioned, with the considerable progress made by pre-training models, multi-modal pre-training has started to attract significant attention from machine learning, computer vision, and natural language processing in recent years, i.e., it has now been a hot interdisciplinary research topic. However, there are still three challenges in large-scale multi-modal pre-training: (1) \textbf{Invalid Strong Assumption}: Most existing models are designed by assuming that there exists strong semantic correlation between the input image-text pair (see Figure~\ref{fig:intro}), but this strong correlation assumption is often invalid in practice. (2) \textbf{Inefficiency of Pre-training}: The pre-training process is often very expensive, and a large number of GPUs are needed for parallel pre-training. (3) \textbf{Difficulty in Model Deployment}: The pre-training models are typically too large to be deployed in real-world application scenarios, which is especially harder for those single-tower models (e.g., UNITER~\cite{chen2020uniter}). In this project, to overcome the above three challenges, we propose a novel two-tower pre-training model called BriVL within the cross-modal contrastive learning framework (like OpenAI CLIP~\cite{radford2021learning}), instead of the single-tower architecture that is adopted by most multi-modal pre-training models. Importantly, unlike OpenAI CLIP~\cite{radford2021learning}, we devise a more advanced cross-modal contrastive learning algorithm based on the latest MoCo~\cite{he2020momentum} so that our BriVL can incorporate \emph{more negative samples in limited GPU resources}. Our motivation for model design is detailed below.

Most existing multi-modal pre-training models, especially those with the single-tower architecture~\cite{li2020oscar,uppal2020emerging,qi2020imagebert,xia2020xgpt,desai2020virtex,gan2020large,yu2020ernie,messina2020fine,cho2020x,gu2020self}, take an assumption that there exists strong semantic correlation between the input image-text pair. With this strong assumption, the interaction between image-text pairs can thus be modeled with cross-modal transformers. However, in real-world application scenarios, the strong correlation assumption is often invalid. For example, there often exists only weak correlation between image-text pairs, as illustrated in Figure~\ref{fig:intro}. Moreover, we also conduct extensive experiments and find that the performance of the two-tower models is significantly better than that of the single-tower models on the noisy image-text data (e.g., crawled from the Web). In this project, we thus choose the two-tower architecture to devise our large-scale multi-modal pre-training model.

\begin{figure}[t]
    \centering
    \includegraphics[width=0.46\textwidth]{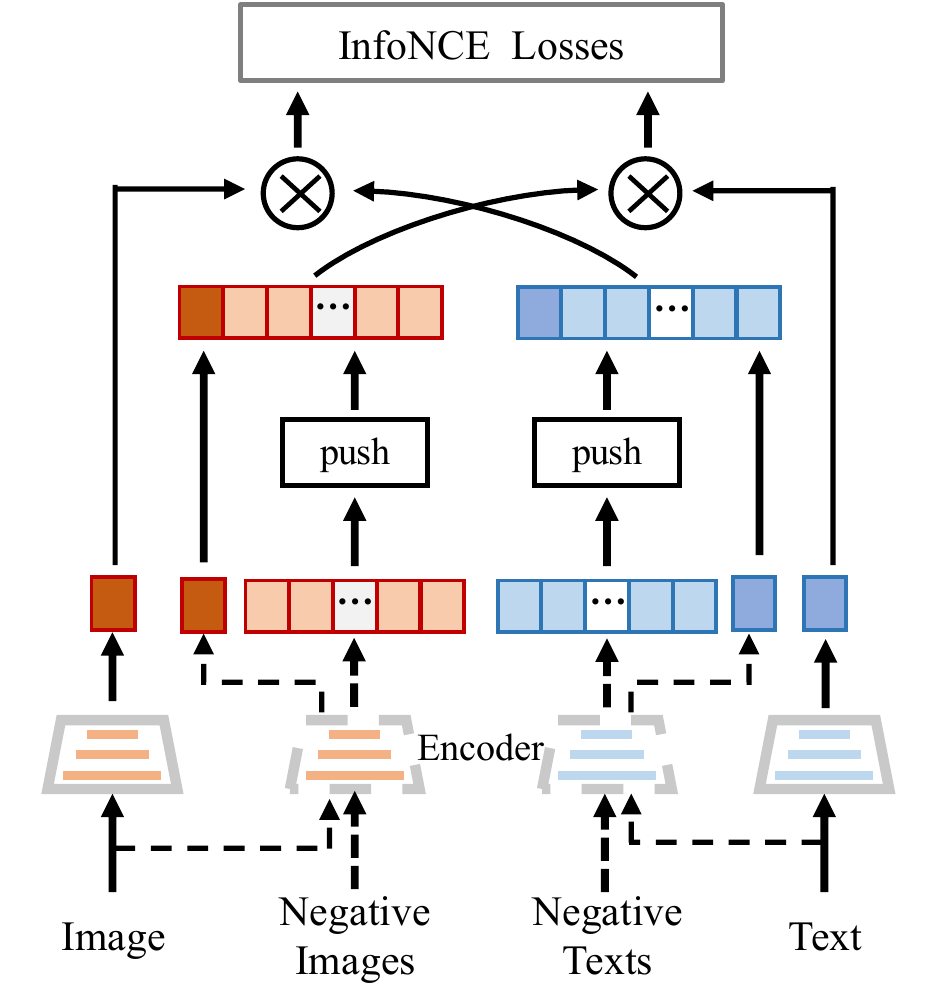}
    \caption{
    A schematic illustration of our BriVL model within the cross-modal contrastive learning framework.
    }
    \label{fig:tower}
    \vspace{-0.15in}
\end{figure}

Specifically, given the web-crawled image-text data for pre-training, we need to design a multi-modal pre-training model based on the two-tower architecture.
However, such network architecture is too simple (without fine-grained cross-modal interaction like UNITER) and its representation ability must be enforced for multi-modal pre-training. Thanks to the recent progress of self-supervised learning~\cite{wu2018unsupervised, oord2018representation, hjelm2019learning, zhuang2019local, bachman2019learning}, contrastive learning~\cite{chen2020simple,NEURIPS2020_BYOL,chen2020exploring,tosh2021contrastive} has been found to significantly improve the representation ability of deep neural networks. Following this idea, we introduce comparative learning into our two-tower architecture. However, unlike OpenAI CLIP~\cite{radford2021learning} that adopts a simple contrastive learning method with the requirement of large batches, we devise a more advanced cross-modal contrastive learning algorithm. As illustrated in Figure~\ref{fig:tower}, given a specific image-text pair, the image modality or the text modality can be used to construct absent samples of the image-text pair, and the number of negative samples is expanded based on the latest MoCo~\cite{he2020momentum} framework to improve the representation ability of the neural network.
By building a large queue-based dictionary, our model can incorporate more negative samples in limited GPU resources, leading to even better results in image-text retrieval.

\begin{figure*}[t]
    \centering
    \includegraphics[width=0.94\textwidth]{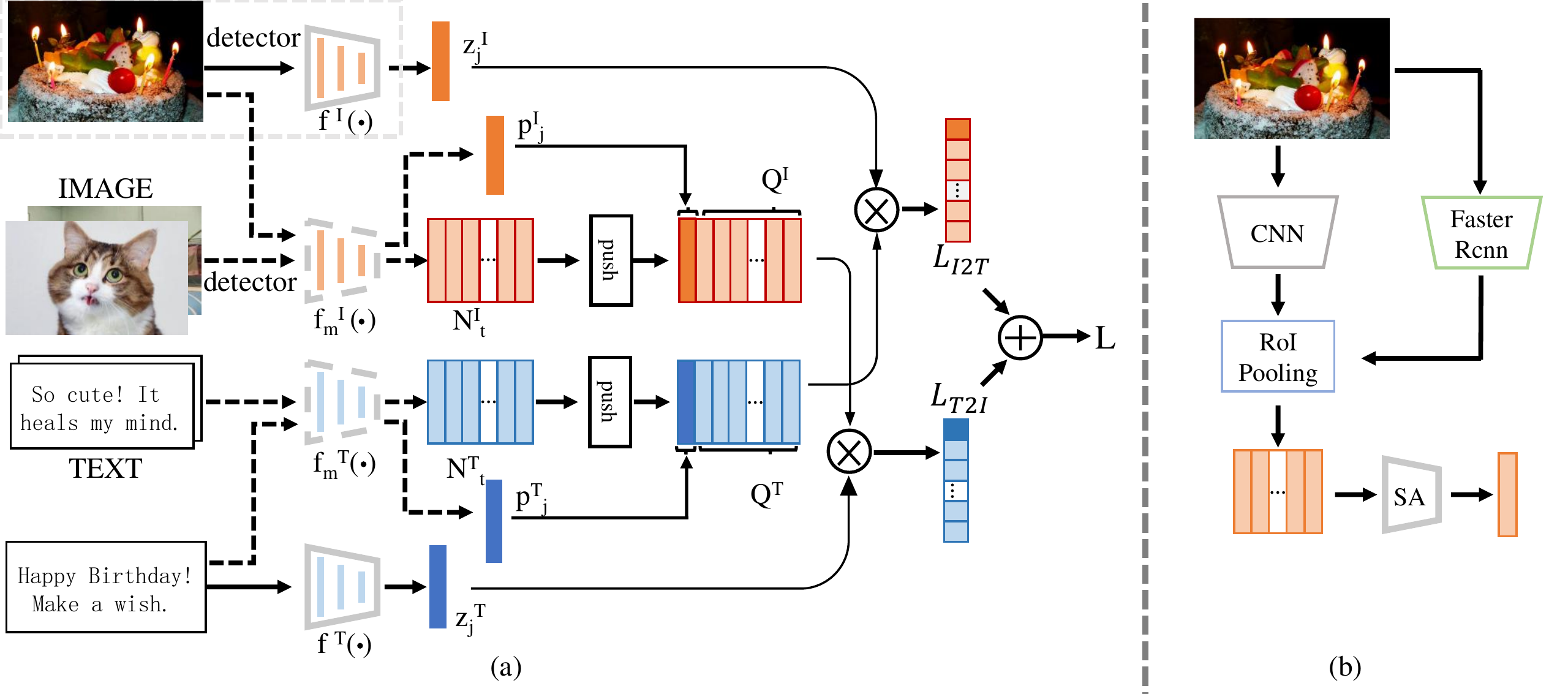}
    \caption{
    (a) A schematic illustration of the proposed BriVL model for large-scale multi-model pre-training. (b) The architecture of the image encoder $f^I$ used for BriVL. Notation: SA -- self-attention based on transformer.
    }
    \label{fig:cross-modal}
    \vspace{-0.1in}
\end{figure*}

Due to the usage of the two-tower architecture as well as the contrastive-learning based pre-training strategy, our proposed BriVL model has a high flexibility and can be readily deployed in real-world application scenarios. It mainly has three advantages:
(i) With a two-tower architecture, the text encoder and the image encoder can be easily replaced with the latest larger single-modal pre-training models, further enforcing the representation ability of our BriVL model.
(ii) Once our BriVL model is pre-trained, it can provide cloud-accessible APIs of the image and text feature embeddings as well as the matching score of an image-text pair, which are very convenient to be deployed in various downstream tasks. Particularly, when a vector engine is used to speed up the inference stage, the efficiency of image-text retrieval can be significantly improved.
(iii) It is convenient to add other pre-training tasks (e.g., image-to-text generation) into our BriVL model. Note that our image-to-text generation (i.e., image captioning) model achieves the new state-of-the-art on the AIC-ICC~\cite{wu2017ai}  dataset.

Our main contributions are three-fold: (1) We have constructed a large Chinese multi-source image-text dataset called RUC-CAS-WenLan for multi-modal pre-training. The first version of RUC-CAS-WenLan consists of 30 million image-text pairs, which come from the rich image-text content generated by web users, including news, sports, entertainment, culture, and other topics. In the near future, this pre-training dataset will be enlarged to 500 million image-text pairs. (2) We have proposed the first large-scale Chinese multi-modal pre-training model called BriVL. The first version of our BriVL model pre-trained on RUC-CAS-WenLan has 1 billion parameters. Importantly, our BriVL model outperforms both UNITER~\cite{chen2020uniter} and OpenAI CLIP~\cite{radford2021learning} on the RUC-CAS-WenLan test set and AIC-ICC~\cite{wu2017ai} validation set. In the near future, our BriVL model will contain 10 billion parameters, which will be pre-trained with 500 million image-text pairs.

\vspace{-0.2cm}
\section{Methodology}
\label{sec:cmp}

Our cross-modal pre-training model is defined based on the image-text retrieval task. Our main goal is thus to learn two encoders that can embed image and text samples into the same space for effective image-text retrieval.
To enforce such cross-modal embedding learning, we introduce contrastive learning with the InfoNCE loss~\cite{oord2018representation} into our BriVL model, as illustrated in Figure~\ref{fig:cross-modal}. Specifically, for a given text embedding, our learning objective is to find the best image embedding from a batch of image embeddings. Similarly, for a given image embedding, our learning objective is to find the best text embedding from a batch of text embeddings.
In one word, our pre-training model learns a cross-modal embedding space by jointly training the image and text encoders to maximize the cosine similarity of the image and text embeddings of the true pair for each sample in the batch while minimizing the cosine similarity of the embeddings of the other incorrect pairs.
This results in an InfoNCE loss over each batch of image-text pairs for pre-training our BriVL model.
Note that our model can incorporate more negative samples in limited GPU resources comparing to OPENAI CLIP, leading to even better results in image-text retrieval (see Section~\ref{sec:expwenlan}).

Formally, for the image-text retrieval task, we denote the training set as $\mathcal{D} = \left\{(x^I_{i}, x^T_{i})| i=1,\cdots,N \right\}$, where $(x^I_{i}, x^T_{i})$ is a matched image-text pair from the RUC-CAS-WenLan dataset, and $N$ is the size of $\mathcal{D}$.
Our image-text retrieval model leverages contrastive learning and expands the latest MoCo~\cite{he2020momentum} as the pre-training framework, as illustrated in Figure~\ref{fig:cross-modal}. Each image $x^I_{i}$ (or each text $x^T_{i}$) is encoded by the image encoder $f^{I}$ (or text encoder $f^{T}$) to obtain its 1-D embedding  $z_{i}^I$ (or $z_{i}^T$). The image encoder (see Figure~\ref{fig:cross-modal}(b)) contains a CNN backbone and a successive self-attention block. A sequence of object embeddings is obtained using a object detector to downsample the feature map from CNN and then encoded by the self-attention block. The text encoder is stacked by several self-attention blocks such as RoBERTa~\cite{liu2019roberta}.
A two-layer Muti-Layer Perception (MLP) block with a RELU activation function is used for mapping each encoder’s representation to the joint cross-modal embedding space.
The parameters of $f^{I}$ and $f^{T}$ are denoted as $\theta^{I}$ and $\theta^{T}$, respectively.

Note that MoCo provides a mechanism of building dynamic dictionaries for contrastive learning, which can be used with various pretext tasks. In this work, we adopt a simple instance discrimination task: a query of an image matches a key of an augmented text if the image corresponds to the text, and vice versa. Further, the introduction of a queue decouples the dictionary size from the mini-batch size. As a result, the dictionary size can be much larger than a typical mini-batch size, and we can set it as a hyper-parameter. Given the momentum parameter $m$, two momentum-updated encoders $f^{I}_m$ (with the parameters $\theta^{I}_m$) and $f^{T}_m$ (with the parameters $\theta^{T}_m$) are kept for the image and text modalities, respectively. Their update rule is given by:
\begin{equation}
\theta^I_m = m\cdot\theta^I_m+(1-m)\cdot\theta^I
\end{equation}
\begin{equation}
\theta^T_m = m\cdot\theta^T_m+(1-m)\cdot\theta^T
\end{equation}

Similar to MoCo, BriVL maintains two queues $Q^I$ and $Q^T$, which contain $K$ image negative keys and $K$ text negative keys, respectively. Given batch size $bs$ in the pre-taining stage, after each iteration, all $bs$ image negative keys and $bs$ text negative keys are separately pushed into these two queues. In this way, keys in queues are updated in each iteration. Specifically, at iteration $t$, the image and text negative keys from the current data batch $\{B^I_t,B^T_t\}$ are calculated by forwarding the momentum-updated encoders $f^{I}_m$ and $f^{T}_m$: $\mathcal{N}_t^I = \{f_m^I(x_j^I)|x_j^I \in B^I_t\}$, and $\mathcal{N}_t^T = \{f_m^T(x_j^T)|x_j^T \in B^T_t\}$. $\mathcal{N}_t^I$ and $\mathcal{N}_t^T$ are then updated to $Q^I$ and $Q^T$, respectively. Moreover, the positive key is unique for each image query $x_j^I$ (or text query $x_j^T$), and it is obtained also by forwarding the momentum-updated encoders: $p_j^T = f_m^T(x_j^T)$ (or $p_j^I = f_m^I(x_j^I)$). The loss function for each data batch is constructed as follows: for each image query $x_j^I$, we define the contrastive loss between its image embedding $z_j^I$ and all positive/negative text keys in the queue $Q^T$, and then obtain an InfoNCE loss:
\begin{equation}
\mathcal{L}_{I2T} = \hspace{-0.03in}  - \hspace{-0.03in} \sum_{j}\log\frac{ \exp(z_j^I\cdot p_j^T / \tau)}{ \exp(z_j^I\cdot p_j^T / \tau) + \hspace{-0.1in}   \sum\limits_{n^T \in Q^T} \hspace{-0.1in} \exp(z_j^I \cdot n^T / \tau)}   
\end{equation}
where $n_T$ denotes a text negative key for each image query and the hyper-parameter $\tau$ denotes the temperature. The similarity is measured by dot product here. Similarly, for each text query $x_j^T$, the InfoNCE loss is formulated as:
\begin{equation}
\mathcal{L}_{T2I} = \hspace{-0.03in}  - \hspace{-0.03in} \sum_{j}\log\frac{ \exp(z_j^T\cdot p_j^I / \tau)}{ \exp(z_j^T\cdot p_j^I / \tau) + \hspace{-0.1in}   \sum\limits_{n_I \in Q^I} \hspace{-0.1in} \exp(z_j^T \cdot n^I / \tau)}   
\end{equation}
where $n_I$ denotes an image negative key of each text query.

The total loss function for BriVL is defined as:
\begin{equation}
\mathcal{L}_{total} = \mathcal{L}_{I2T} + \mathcal{L}_{T2I}
\end{equation}
In the test/evaluation stage, the query image (or text) is also retrieved simply by the dot product defined over the output (i.e., embeddings) of the pre-trained encoders.

Due to its high flexibility, our BriVL model can be readily deployed in a wide range of application scenarios. First, other pre-training tasks (e.g. image-to-text generation) can be added to our BriVL model by sharing the same text or image encoder. Second, the pre-trained text and image encoders can be directly applied to many downstream multi-modal tasks such as image-to-text retrieval, text-to-image retrieval, text-to-image generation \cite{reed2016generative} and visual dialog \cite{niu2019recursive}. This actually leads to several downstream applications developed based on our BriVL model.

\section{Experiments}

\subsection{Dataset and Settings}

\noindent\textbf{Pre-Training Dataset}~~Our BriVL model is pre-trained on a Web-crawled multi-source image-text dataset. This dataset is part of the WenLan project, called \emph{RUC-CAS-WenLan} for short. RUC-CAS-WenLan collects image-text pairs from multiple information sources on the Web, including news, encyclopedia (i.e., Baidu Baike) and Weibo. Images from these data sources are selected to form image-text pairs together with their corresponding text descriptions. Since the obtained image-text pairs are crawled from the Web, there exist much noise in the original data. Thus, we then perform an elaborate cleaning process (e.g., duplicate and sensitive information detection) to filter out sensitive or low-quality pairs. For each data source, we also employ topic models to analyze the overall topic distribution and extract topic words, which help select and keep high-quality content information.
Finally, our dataset has kept 30 million image-text pairs covering a variety of topics and content categories, including news, art, education, sports, entertainment, games, and culture. Out of them, 11,000 pairs are randomly selected to form the test set.

\noindent\textbf{Text Encoder}~~As mentioned in Section~\ref{sec:cmp}, a text encoder consists of a textual backbone, a self-attention block, and a two-layer MLP. We choose the encoder of Chinese RoBERTa\_Large\footnote{https://github.com/brightmart/roberta\_zh} as our textual backbone. Note that RoBERTa\_Large includes a total of 24 transformer layers with 1,024 hidden units and 16 heads. The self-attention block consists of 4 transformer layers, designed for capturing the relationships across the textual tokens. The two-layer MLP is used to project the textual embedding to the cross-modal embedding space.

\noindent\textbf{Image Encoder}~~Following UNITER~\cite{chen2020uniter}, we first employ pre-trained Faster-RCNN~\cite{ren2016faster} to detect object bounding-boxes from each image. We further utilize EfficientNet\_B7~\cite{tan2019efficientnet} to extract the visual features of each image for computation efficiency. By applying RoI pooling~\cite{girshick2015fast} on the output of EfficientNet\_B7, we obtain the features of multiple objects and then combine them with a self-attention block (of 4 transformer layers). The fused object features are fed into a two-layer MLP and projected to the cross-modal embedding space.

\noindent\textbf{Implementation Details}~~We utilize the momentum-updated history queue as in MoCo~\cite{he2020momentum} for contrastive learning.
We adopt clip-wise random crops, horizontal flips, Gaussian blur, graying, and color jittering for data augmentation over input images. A non-linear projection head is attached to the text/image encoder to obtain feature vectors in the same size 2,560. Our BriVL model is trained with 15 epochs. We select hyper-parameters heuristically due to computational constraint: the learnable temperature parameter $\tau = 0.05$, momentum $m = 0.99$, and the queue size is 16,384. We adopt the Adam optimizer with decoupled weight decay regularization over all weights that are not gains or biases, and decay the learning rate using a cosine schedule. We use a mini-batch size of 128 for each of the 16 machines (each machine has 8 A100 GPUs), resulting in a total batch size of 2,048. The mixed-precision and half-precision Adam statistics are used to accelerate the pre-training process and save the memory. It takes 7 days to pre-train our BriVL model on 128 A100 GPUs.

\subsection{Results on AIC-ICC}

We select the AIC-ICC caption competition~\cite{wu2017ai} to evaluate our pre-trained BriVL model because it is the only publicly-available Chinese multi-modal dataset. This Chinese caption dataset (called as AIC-ICC) includes about 300,000 images, with 5 candidate Chinese caption texts per image. The validation split (with 30,000 images) of this dataset is used for performance evaluation on two downstream tasks (i.e., image-text retrieval and image captioning). To make a comparison with OpenAI CLIP on this dataset, we have to translate the Chinese captions in the validation split into the English ones (with Google Translation). It is noticeable that we can only obtain the inference code\footnote{https://github.com/openai/CLIP} (but not the training code) of CLIP from OpenAI, and thus are unable to pre-train CLIP on our own RUC-CAS-WenLan dataset.

\begin{table}[t]
    \centering
    \caption{Evaluation results for the text-image retrieval downstream task on the AIC-ICC validation set. }
    \vspace{0.03in}
    \label{tab:retrieval}
    \footnotesize
    \scalebox{0.99}{
    \begin{tabular}{l|c|c|c|c|c|c}
    	\toprule
    	Tasks & \multicolumn{3}{c}{Image-to-Text Retrieval} & \multicolumn{3}{|c}{Text-to-Image Retrieval} \\
    	\hline
    	Metrics & R@1 & R@5 & R@10 & R@1 & R@5 & R@10 \\
    	\hline
    	CLIP~\cite{radford2021learning} & 13.4 & 27.3 & 35.1 & 7.8 & 18.5 & 25.0 \\
    	UNITER~\cite{chen2020uniter} & 14.8 & 29.8 & 37.9 & 9.8 & 23.3 & 31.4 \\
    	BriVL~(ours) & \textbf{20.3}	&\textbf{37.0} &\textbf{45.6} &\textbf{14.4} &\textbf{30.4} &\textbf{39.1}\\
    	\bottomrule
    \end{tabular}
    }
\end{table}

\begin{table}[t]
    \centering
    \caption{Evaluation results for the image captioning downstream task on the AIC-ICC validation set. $^*$ denotes the result obtained on the test set. }
    \vspace{0.03in}
    \label{tab:caption}
    \footnotesize
    \scalebox{0.99}{
    \tabcolsep7pt
    \begin{tabular}{l|c|c|c|c}
    	\toprule
    	 Metrics & BLEU & METEOR & ROUGE-L & CIDEr \\
    	\hline
    	CHAMPION'17$^*$  & 62.8 & 43.0 & -- & 210.4  \\
    \hline
    	UNITER~\cite{chen2020uniter} & 62.8 & 38.7 & 69.2 & 199.7  \\
    	BriVL~(ours)  &  \textbf{66.1} & \textbf{41.1} & \textbf{71.9} & \textbf{220.7} \\
    	\bottomrule
    \end{tabular}
    }
\end{table}

Table~\ref{tab:retrieval} presents the image-text retrieval results. We directly leverage the extracted features for nearest-neighbour (NN) retrieval without fine-tuning. We can observe that our BriVL significantly outperforms CLIP and UNITER on both the text-to-image and image-to-text retrieval subtasks, showing the effectiveness of the proposed BriVL in multi-modal pre-training. Note that our BriVL runs about 20 times faster than UNITER (but as fast as CLIP).

Table~\ref{tab:caption} presents the image captioning results. Fine-tuning is conducted on the training split of AIC-ICC. We adopt four widely-used evaluation metrics: BLEU, METEOR, ROUGE-L, and CIDEr. It can be clearly seen that our BriVL performs better than the competitors in terms of three of the four metrics, i.e., our BriVL achieves the best overall performance on the AIC-ICC dataset. This means that our BriVL model also has a good generalization ability in the image captioning downstream task.

\begin{table}[t]
    \centering
    \caption{Evaluation results for the text-image retrieval downstream task on the RUC-CAS-WenLan test set.}
    \vspace{0.03in}
    \label{tab:wenlan}
    \footnotesize
    \scalebox{0.99}{
    \begin{tabular}{l|c|c|c|c|c|c}
    	\toprule
    	Tasks & \multicolumn{3}{c}{Image-to-Text Retrieval} & \multicolumn{3}{|c}{Text-to-Image Retrieval} \\
    	\hline
    	Metrics & R@1 & R@5 & R@10 & R@1 & R@5 & R@10 \\
    	\hline
    	 CLIP~\cite{radford2021learning} & 7.3 & 15.0 & 19.0 & 7.8 & 15.9 & 19.9 \\
    	UNITER~\cite{chen2020uniter}  & 5.3 & 16.9  & 24.6 & 5.7 & 16.7 & 24.3 \\
    	BriVL~(ours)  & \textbf{36.1} & \textbf{55.5}  & \textbf{62.2} & \textbf{36.0} & \textbf{55.4} & \textbf{62.1}\\
    	\bottomrule
    \end{tabular}
    }
\end{table}

\subsection{Results on RUC-CAS-WenLan}
\label{sec:expwenlan}

We further make performance evaluation on the text-image retrieval task on the test split of RUC-CAS-WenLan, which includes 11,000 image-text pairs. Table~\ref{tab:wenlan} presents the text-image retrieval results on the RUC-CAS-WenLan test set. It is noticeable that our BriVL achieves significant improvements over UNITER and OpenAI~CLIP\footnote{The inference code of OpenAI CLIP is directly implemented on the translated test split with Google Translation.}. Particularly, our BriVL leads to more than 45\% performance gaps in terms of R@10 on both retrieval subtasks. This demonstrates the largest advantage of our BriVL in multi-modal pre-training. Furthermore, our BriVL is pre-trained by using 128 GPUs for about 7 days, comparing to OpenAI CLIP using 256 GPUs for 12 days.

\subsection{User Study Results}

\begin{table}[t]
    \centering
    \caption{User study results for the text-image retrieval downstream task. Three human annotators are involved in such user study. }
    \vspace{0.03in}
    \label{tab:user_study}
    \footnotesize
    \scalebox{0.99}{
    \tabcolsep4pt
    \begin{tabular}{l|c|c|c|c}
    	\toprule
    	Tasks & \multicolumn{4}{c}{Image-to-Text Retrieval}  \\
    	\hline
    	Metrics & NDCG@5 & NDCG@10 & NDCG@20 & MAP    \\
    	\hline
        CLIP~\cite{radford2021learning} & 32.9 & 38.8 & 53.0 & 30.3\\
        BriVL  & \bf37.5 & 42.8 & 55.5 & \bf38.3 \\
    	BriVL+UNITER & 37.0 & \bf43.5 & \bf56.3 & 37.6\\

        \hline
        Tasks & \multicolumn{4}{|c}{Text-to-Image Retrieval} \\
        \hline
        Metrics & NDCG@5 & NDCG@10 & NDCG@20 & MAP    \\
    	\hline
    	CLIP~\cite{radford2021learning} & 28.0 & 32.3 & 43.7 & 16.7 \\
    	BriVL & 46.9 & 51.5 & 61.6 & 47.2 \\
    	BriVL+UNITER  & \bf49.9 & \bf55.0 & \bf65.1 & \bf52.5\\

    	\bottomrule
    \end{tabular}
    }
\end{table}

\begin{figure}[t]
    \centering
    \includegraphics[width=0.47\textwidth]{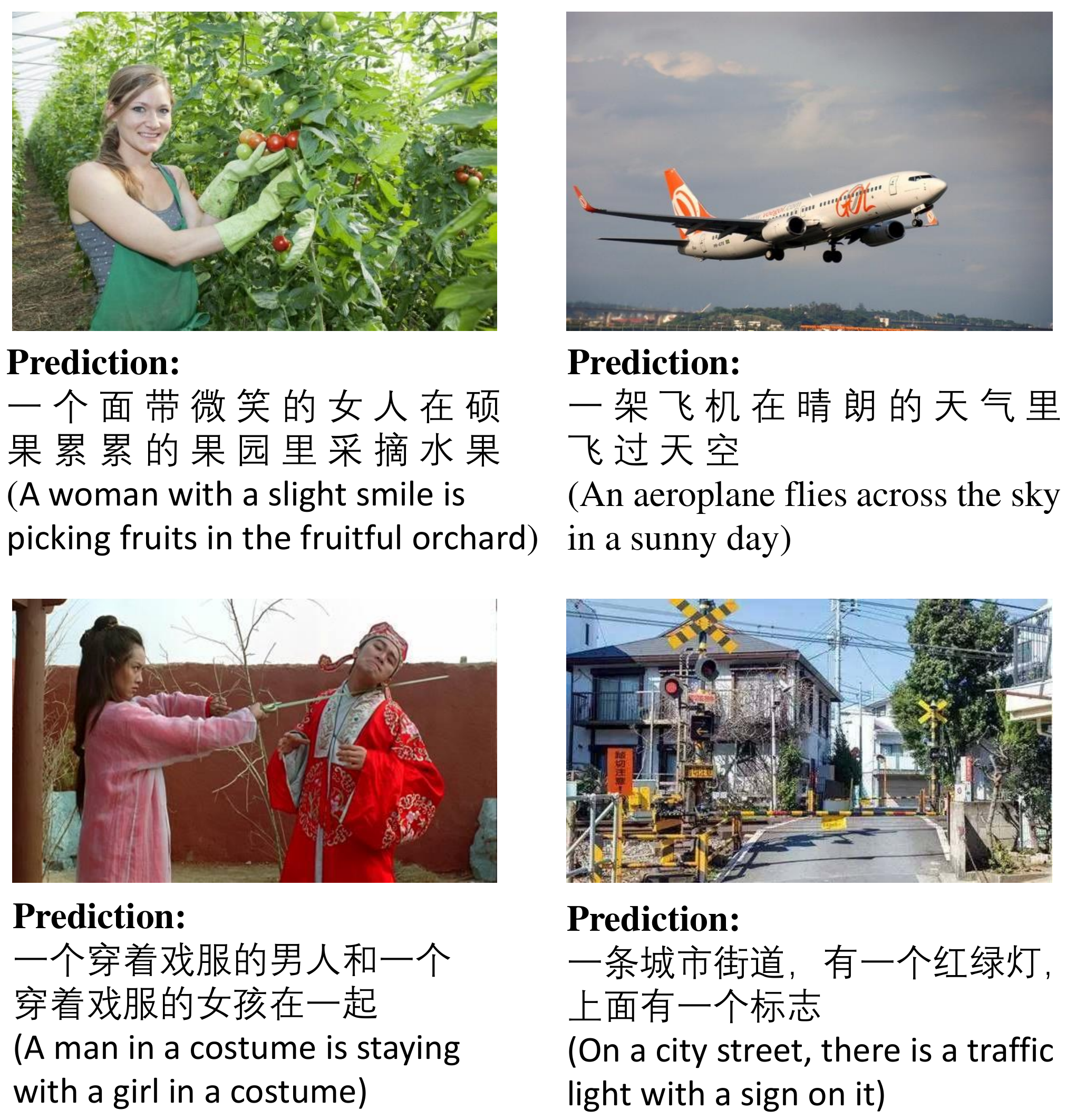}
    \caption{
    Visualization examples obtained by our image captioning model. Note that two data sources (caption and web) are used in the top and bottom rows, respectively. }
    \label{fig:exp_caption}
\end{figure}

The user study is carried out over the text-image retrieval results obtained by the pre-training models (e.g., CMLC and CLIP~\cite{radford2021learning}). We select a group of image and text queries for testing. For each text (or image) query, we retrieve the first 30 results with the tested model from the specified candidate set, and manually score each of the 30 results by 3 ratings (i.e., 0, 1, and 2). Note that the higher the score is, the stronger the correlation between the image and text is. Since three human annotators are involved independently, the final score for each of the 30 results is obtained with 7 ratings (0-6). The scores of each text (or image) query are thus formed into a 30-length score sequence.

The NDCG and MAP metrics are used to evaluate the human retrieval quality. Note that these metrics are widely used for evaluating retrieval quality. Particularly, during computing MAP, the text-image pair is considered to be relevant if the corresponding score is higher than 2. The obtained comparative results are presented in Table~\ref{tab:user_study}. As expected, the user study does validate that our BriVL outperforms OpenAI CLIP~\cite{radford2021learning}. When the candidate set (per query) of UNITER is obtained using our BriVL, UNITER is shown to lead to further improvements over our BriVL (see BriVL+UNITER vs. BriVL).

\begin{figure}[t]
    \centering
    \includegraphics[width=0.47\textwidth]{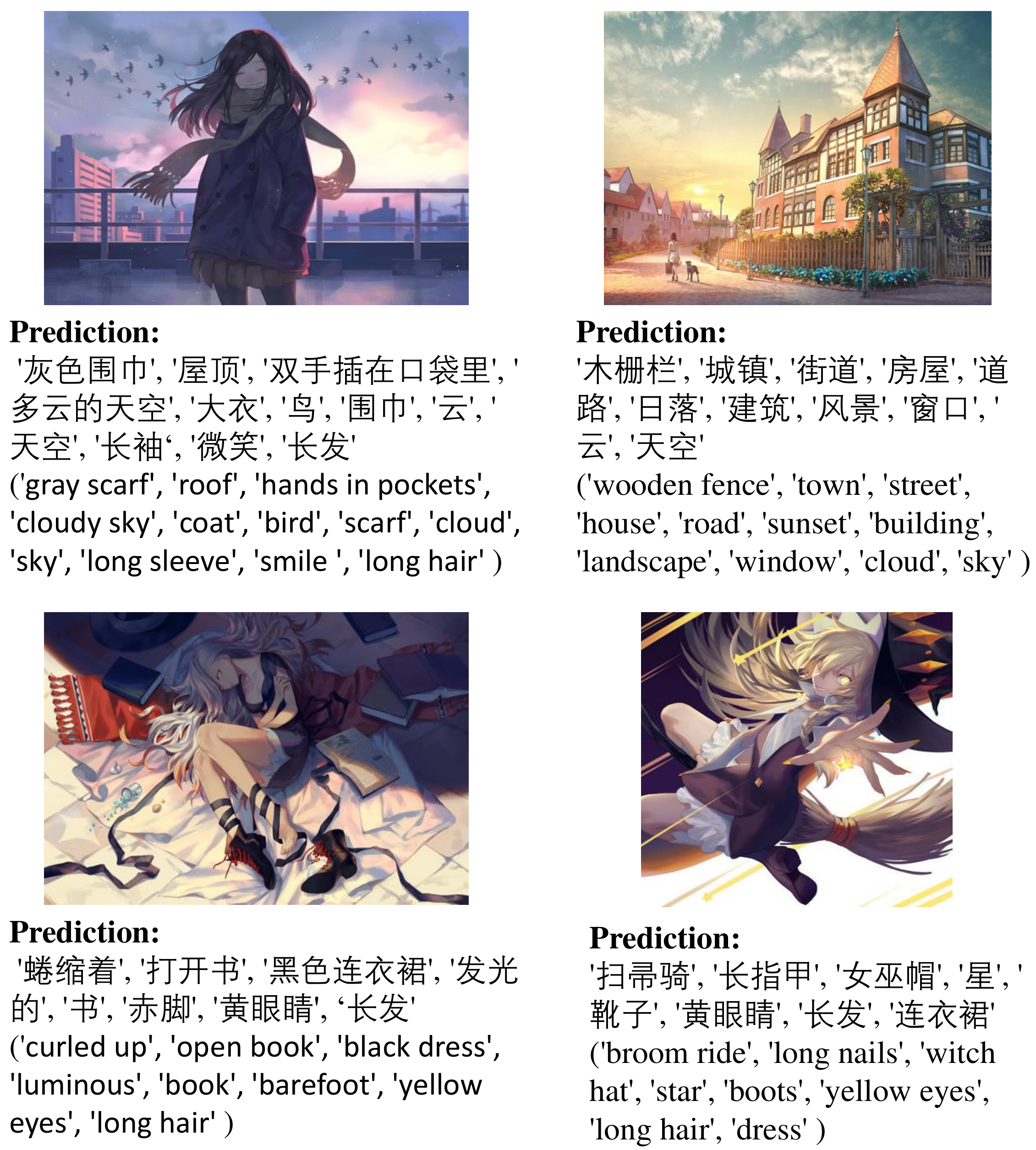}
    \caption{
    Visualization examples obtained by our image tagging model. Note that our image tagging model is almost the same as our image captioning model. The anime images are used in this downstream task.}
    \label{fig:exp_tagging}
\end{figure}

\subsection{Visual Results}

Figure~\ref{fig:exp_caption} presents the visualization  examples obtained by our image captioning model. Note that two data sources (caption and web) are used in the top and bottom rows, respectively. We can observe that the generated captions by our model are fluent, vivid, and accurate to express the semantic meanings of the input pictures. This suggests that multi-model pre-training indeed brings benefits to the image captioning downstream task.

Figure~\ref{fig:exp_tagging} presents the visualization examples obtained by our image tagging model. Note that our image tagging model is almost the same as our image captioning model. The anime images are used in this downstream task. We can see that our image tagging model is able to predict accurate tags for each anime image. This provides evidence that multi-model pre-training indeed brings benefits to the image tagging downstream task.

\section{Downstream Applications}

Although `WenLan' can be applied to a variety of cross-modal downstream tasks, we have only developed two web applications, MatchSoul and Soul-Music, at this moment. Our main goal is to directly demonstrate the power of multi-modal pre-training in real-world scenarios. We will develop more applications in the near future.

\subsection{MatchSoul}

MatchSoul is developed based on our pre-trained BriVL model. Note that we directly deploy our pre-trained model without any fine-tuning. This application is devised as follows: given a picture uploaded by user, it returns an `golden' sentence that is the most relevant to this picture.

Unlike the general image generation, this application does not generate a descriptive sentence for the input picture. In contrast, it chooses to match the picture with the `golden' sentence (from a candidate set of 300,000 `golden' sentences) according to the characteristics of the picture, as illustrated in Figure~\ref{fig:exp_application}(a). The chosen `golden' sentences are humor, literary, and philosophical thinking. We look forward to giving users a sense of surprise and playing the finishing touch to the picture.

\subsection{Soul-Music}

\begin{figure}[t]
    \centering
    \includegraphics[width=0.47\textwidth]{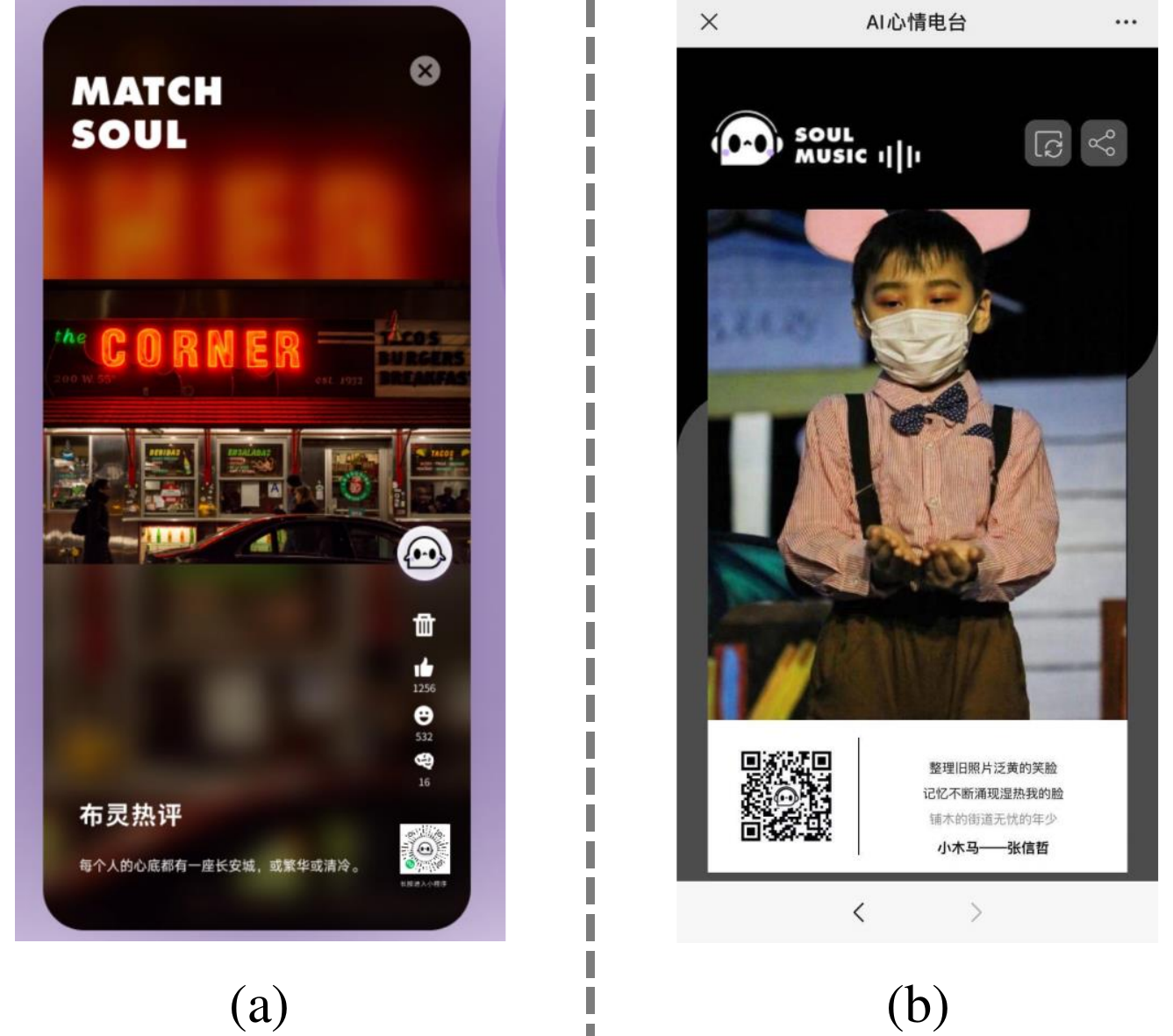}
    \caption{
    Demonstration of our downstream application. (a) MatchSoul: matching pictures with `golden' sentences. (b) Soul-Music: matching pictures with `golden' lyrics. }
    \label{fig:exp_application}
\end{figure}

Similar to MatchSoul, Soul-Music is also developed based on our pre-trained BriVL model. Specifically, given a picture uploaded by user, Soul-music returns a song lyric that well fits the artistic conception of this picture. As illustrated in Figure~\ref{fig:exp_application}(b), Soul-Music matches the input picture with the most relevant song lyric, and even accurately localizes a part of the song lyric which best matches the characteristics of this picture.

\section{Conclusion and Future Work}

This paper presents the first large-scale Chinese multi-modal pre-training model called BriVL. The first version of our BriVL model has 1 billion parameters, which is pre-trained on the RUC-CAS-WenLan dataset with 30 million image-text pairs. As a part of this project, RUC-CAS-WenLan is a large Chinese multi-source image-text dataset constructed by ourselves for multi-modal pre-training. It is noticeable that our BriVL model significantly outperforms both UNITER and OpenAI CLIP on the RUC-CAS-WenLan test set and AIC-ICC validation set. With the pre-trained BriVL model, we have also developed two web applications called MatchSoul and Soul-Music. In the near future, our BriVL model will be enlarged to 10 billion parameters, which will be pre-trained with 500 million image-text pairs. Moreover, we will also exploit the text-to-image generation pretext task for multi-modal pre-training.

\subsection*{Acknowledgements}

This work was supported in part by National Natural Science Foundation of China (61976220 and 61832017), Beijing Outstanding Young Scientist Program (BJJWZYJH012019100020098), and Large-Scale Pre-Training Program of Beijing Academy of Artificial Intelligence (BAAI).

{
\bibliographystyle{ieee_fullname}
\bibliography{wenlan}
}

\end{document}